\documentclass[10pt,twocolumn,letterpaper]{article}

\usepackage{wacv}
\usepackage{times}
\usepackage{epsfig}
\usepackage{graphicx}
\usepackage{amsmath, colortbl}
\usepackage{amssymb}
\usepackage{booktabs}
\usepackage{enumitem}
\usepackage{xcolor}
\usepackage[accsupp]{axessibility}  
\setlist{nolistsep}

\def\wacvPaperID{596} 

\wacvalgorithmstrack   

\wacvfinalcopy 

\ifwacvfinal
\usepackage[breaklinks=true,bookmarks=false]{hyperref}
\else
\usepackage[pagebackref=true,breaklinks=true,colorlinks,bookmarks=false]{hyperref}
\fi

\begin{document}

\def\Approach{PSENet}
\title{\Approach: Progressive Self-Enhancement Network \\for Unsupervised Extreme-Light Image Enhancement}

\author{Hue Nguyen \qquad Diep Tran \qquad Khoi Nguyen \qquad Rang Nguyen\\
VinAI Research, Vietnam\\
{\tt\small \{v.huent88, v.diepttn147, v.khoindm, v.rangnhm\}@vinai.io}
}

\maketitle
\thispagestyle{empty}

\begin{abstract}
   The extremes of lighting (e.g. too much or too little light) usually cause many troubles for machine and human vision. Many recent works have mainly focused on under-exposure cases where images are often captured in low-light conditions (e.g. nighttime) and achieved promising results for enhancing the quality of images. However, they are inferior to handling images under over-exposure. To mitigate this limitation, we propose a novel unsupervised enhancement framework which is robust against various lighting conditions while does not require any well-exposed images to serve as the ground-truths. Our main concept is to construct pseudo-ground-truth images synthesized from multiple source images that simulate all potential exposure scenarios to train the enhancement network. Our extensive experiments show that the proposed approach consistently outperforms the current state-of-the-art unsupervised counterparts in several public datasets in terms of both quantitative metrics and qualitative results.  Our code is available at \href{https://github.com/VinAIResearch/PSENet-Image-Enhancement}{https://github.com/VinAIResearch/PSENet-Image-Enhancement}.
\end{abstract}

\section{Introduction}
Producing images with high contrast, vivid color, and rich details is one of the important goals of photography. However, acquiring such pleasing images is not always a trivial task due to harsh lighting conditions, including extreme low lighting or unbalanced lighting conditions caused by backlighting. The resulting under-/over-exposed images usually decrease not only human satisfaction but also computer vision system performance on several downstream tasks such as object detection \cite{Rashed_2019} or image segmentation~\cite{tan2020nighttime}. Wrong exposure problems occur early in the capturing process and are difficult to fix once the final 8-bit image has been rendered. This is because the in-camera image signal processors usually use highly nonlinear operations to generate the final 8-bit standard RGB image~\cite{nguyen2016raw, karaimer2016software, nguyen2018raw}. 

\begin{figure}[t]
\centering
\includegraphics[width=0.47\textwidth, page=1]{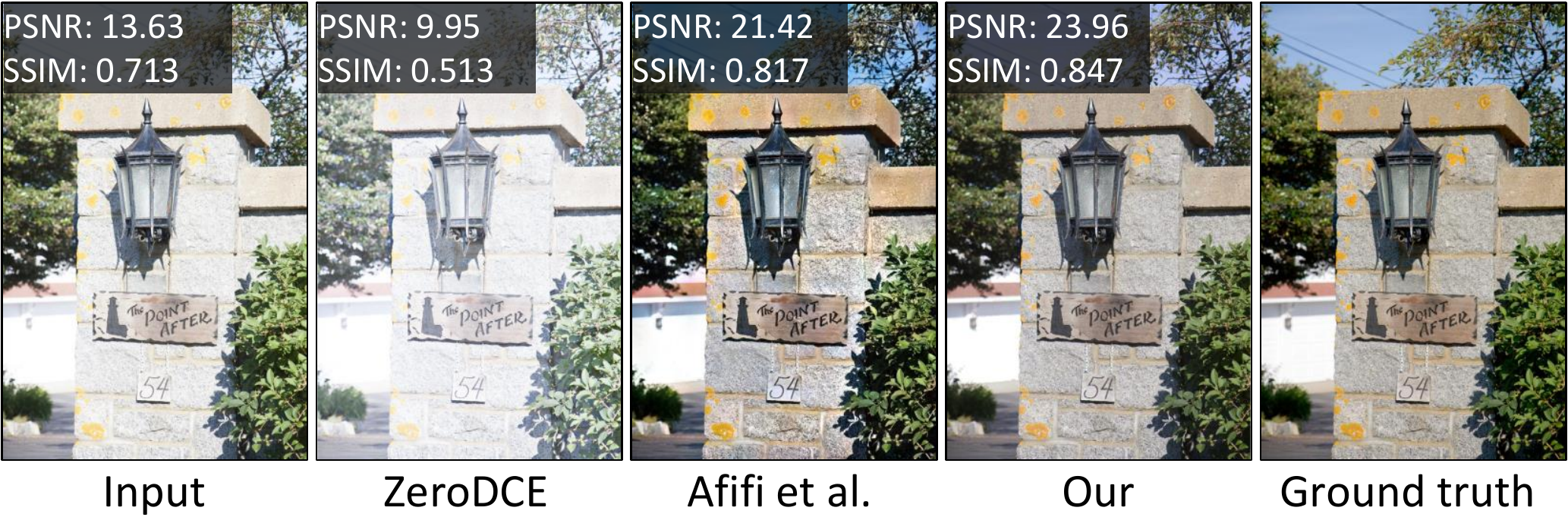}
\caption{Visual comparison on an over-exposed scene. Most of the previous state-of-the-art failed to recover the over-exposed case except recent work proposed by Afifi \etal{}~\cite{afifi2021learning}, which is trained with full supervision.}
\label{fig:teaser}
\end{figure}

Many recent works have mainly focused on under-exposure cases where images are often captured in low-light conditions (\eg{} nighttime). These works have achieved promising results for enhancing the quality of images even captured under extreme low-light conditions. 
However, they failed to handle over-exposure images, as shown in Fig.~\ref{fig:teaser}. The recent work proposed by Afifi \etal{}~\cite{afifi2021learning} achieves impressive results in improving both under- and over-exposed cases. However, their proposed method is designed to work in a supervised manner, requiring a large dataset of wrongly exposed and corresponding ground-truth (GT) well-exposed image pairs. This data collection is typically time-consuming and expensive.

In this paper, we propose a novel unsupervised approach that does not require any well-exposed GT images. The key idea is to generate a pseudo GT image given the input wrongly exposed one in order to train an enhancement network. The pseudo GT image across training epochs is progressively generated by choosing the visually best regions taken from multiple sources, namely the output of the same input image from the previous epoch, the brighter/darker reference images by changing the gamma value of the input image, and the input image itself. The choosing criteria are well-exposedness, local contrast, and color saturation, which are driven by human knowledge of a visually good image and have been shown to be effective in measuring perceptual image quality~\cite{exposure_fusion}. In this way, the task of generating pseudo GT images is simply comparing and selecting the best regions from different sources where almost possible cases of exposure are simulated in training. Furthermore, by using the output of the previous epoch as a source for choosing, we ensure that the output of the current epoch will be better than or at least equal to that of the previous one, giving the name of our approach \Approach~-- Progressive Self Enhancement Network.
 
Our contributions are summarized as follows:
\begin{itemize}
    \item We introduce a new method for generating effective pseudo-GT images from given wrongly-exposed images. The generating process is driven by a new non-reference score reflecting the human evaluation of a visually good image. 
    
    \item We propose a novel unsupervised progressive pseudo-GT-based approach that is robust to various severe lighting conditions, i.e. under-exposure and over-exposure. As a result, the burden of gathering the matched image pairs is removed. 
    
    \item Comprehensive experiments are conducted to show that our approach outperforms previous unsupervised methods by large margins on the SICE \cite{sice} and Afifi \cite{afifi2021learning} datasets and obtains comparable results with supervised counterparts.
    
\end{itemize}

\section{Related Work}
Image enhancement approaches can be divided into two categories: traditional and learning-based methods.

\noindent\textbf{Traditional methods.} One of the simplest and fastest approaches is to transform single pixels of an input image by a mathematical function such as linear function, gamma function, or logarithmic function.  For example, histogram equalization-based algorithms stretch out the image's intensity range using the cumulative distribution function, resulting in the image's increased global contrast. The Retinex theory~\cite{retinexTheoryLand}, on the other hand, argues that an image is made from two components: reflectance and illumination. By estimating the illumination component of an image, the dynamic range of the image can be easily adjusted to reproduce images with better color contrast. However, most Retinex algorithms use Gaussian convolution to estimate illumination, thus leading to blurring edges \cite{exprimentbasedreview}. Frequency-domain-based methods, by contrast, preserve edges by employing the high-pass filter to enhance the high-frequency components in the Fourier transform domain \cite{homomorphicFilter}. However, the adaptability of such traditional methods is often limited due to their unawareness of the overall and local complex gray distribution of an image \cite{exprimentbasedreview}. For a systematic review of conventional approaches, we suggest readers refer to the work of Wang \etal{} \cite{exprimentbasedreview}. 

\noindent\textbf{Learning-based methods.} In recent years, there has been increasing attention to learning-based photo-enhancing methods in both supervised and unsupervised manners.

\noindent\textit{Supervised learning} methods aim to recover natural images by either directly outputting the high quality images ~\cite{lore2016llnet,Lv2018MBLLENLI,Li2020LuminanceawarePN,Xu2020LearningTR} or learning specific parameters of a parametric model (\eg{} Retinex model)~\cite{hdrnet,DeepUPE,moran2020deeplpf} from a paired dataset.
SID~\cite{SeeInTheDark} is a typical example in the first direction. In this work, the authors collect a short-exposure low-light image dataset and adopt a vanilla Unet architecture \cite{ronneberger2015unet} to produce an enhanced sRGB image from raw data thus replacing the traditional image processing pipeline. Following this work, Lamba and Mitra~\cite{RealTimeDarkImageRestorationCvpr2021} present a novel network architecture that concurrently processes all the scales of an image and can reduce the latency times by 30\% without decreasing the image quality. Different from the previously mentioned approaches, Cai \etal{}~\cite{sice} explore a new direction in which both under and over-exposed images are considered. They introduce a novel two-stage framework trained on their own multi-exposure image dataset, which enhances the low-frequency and high-frequency components separately before refining the whole image in the second stage. Afifi \etal{}~\cite{afifi2021learning} put a further step in this direction by introducing a larger dataset along with a coarse-to-fine neural network to enhance image qualities in both under- and over-exposure cases. For learning a parametric model, Retinex theory~\cite{Land1971LightnessAR} is often adopted~\cite{retinexnet,Li2018LightenNetAC,DeepUPE}. Benefiting from paired data, the authors focus on designing networks to estimate the reflectance and illumination of an input image. Dealing with the image enhancement task differently, HDRNet~\cite{hdrnet} presents a novel convolutional neural network to predict the coefficients of a locally-affine model in bilateral space using pairs of input/output images.
\noindent\textit{Unsupervised learning.} Collecting paired training data is always time-consuming and expensive. To address this issue, an unpaired GAN-based method named EnlightenGAN is proposed in~\cite{enlightengan}. The network, including an attention-guided U-Net as a generator and global-local discriminators, shows promising results even though the corresponding ground truth image is absent. To further reduce the cost of collecting reference ground truth images, a set of methods~\cite{Zhang2019ZeroShotRO,Zhu2020ZeroShotRO,zerodce,dce_extension} that do not require paired or unpaired training data are proposed. Two recent methods in this category named ZeroDCE~\cite{zerodce} and Zheng and Gupta~\cite{semanticguided} show impressive results in low-light image enhancement tasks by using a CNN model trained under a set of no reference loss functions to learn an image-specific curve for producing a high-quality output image. However, these methods seem to perform poorly when extending to correct over-exposed images, as shown in Fig.~\ref{fig:teaser}.

Our proposed method, in contrast, is the first deep learning work handling these extreme lighting conditions in an unsupervised manner.

\section{Methodology}

\begin{figure*}[t]
\centering
\includegraphics[width=\textwidth]{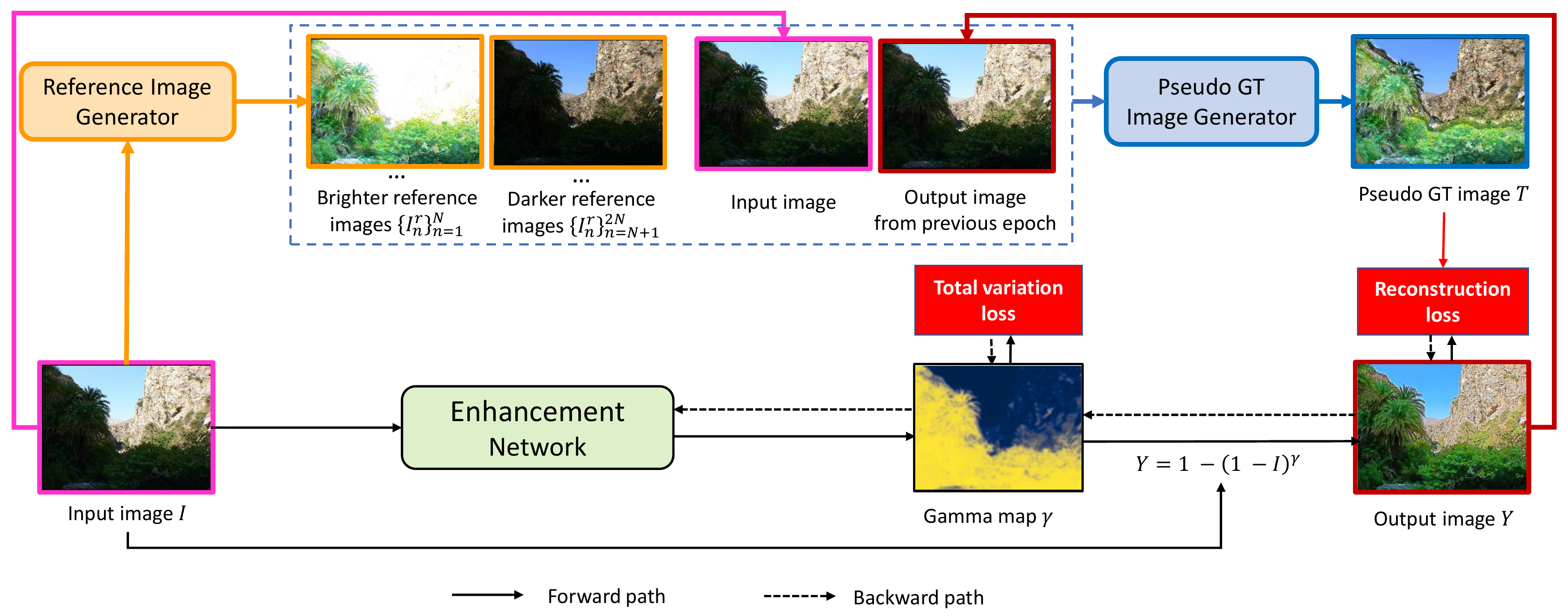}
\caption{Overview of our proposed framework which comprises three main modules: reference image generator, pseudo GT image generator, and enhancement network. Given the input image $I$, the reference image generator randomly generates multiple reference images with different exposure values, half of them are brighter than the input image while the rest are darker. The pseudo GT image generator then takes the input image, the output image from the previous epoch, and the generated reference images as input to produce the pseudo GT image $T$ which is visually better than each of the input components alone based on our proposed non-reference scoring criteria. Finally, the enhancement network predicts the gamma map $\gamma$ to transform the original image $I$ to obtain the output image $Y$. The enhancement network is trained with two loss functions: reconstruction loss between the output image $Y$ and the pseudo GT image $T$, and the total variation loss applied on the predicted gamma map $\gamma$ to imply the smoothness in prediction. It is worth noting that only the enhancement network is used in testing.}
\label{fig_framework-overview}
\end{figure*}

Given an sRGB image, $I$, captured under a harsh lighting condition with low contrast and washed-out color, our approach aims to reconstruct the corresponding enhanced image $Y$, which is visibly better and visually pleasant in terms of contrast and color without any supervision. 

To address the problem, our key contribution is to propose a new self-supervised learning strategy for training the image enhancement network. That is, we randomly synthesize a set of reference images to be combined together to produce a synthetically high-quality GT image for training. The way of combination is driven by the human knowledge of how visually good an image is. To our best knowledge, our \textit{unsupervised} method is the first to produce pseudo-GT images for training on a large set of ill-exposed images; while other data synthesized methods use well-exposed images as GT to generate corresponding ill-exposed inputs. By using this approach, our model does not suffer from the domain gap issue. Compared with image fusion, which only produces a single output image for input, our pseudo GT images are progressively improved after each epoch, making our model adapt to a wide range of lighting conditions  (see Sec. \ref{sec_experiments} for empirical evidence).

In detail, our reference image generator first takes an image as input and generates $2N$ images where the first $N$ images are darker and the rest are brighter compared to the original input image. Then, the pseudo GT generator module uses these reference images along with the input and the previous prediction of the enhancement network to create the pseudo GT image. It is worth noting that including the previous prediction in the set of references ensures that the quality of the pseudo GT image is greater or at least equal to the previous prediction according to our proposed non-reference score, thus making our training progressively improved. Our training framework is illustrated in Fig.~\ref{fig_framework-overview} and the detail of each module will be described in the following sections.

\subsection{Random Reference Image Generation} \label{refence_generator}
\begin{figure}[t]
\centering
\includegraphics[width=0.45\textwidth, page=2]{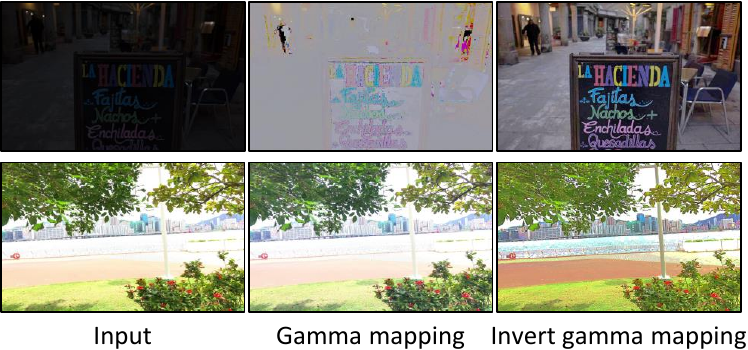}
\caption{Outputs of the gamma and invert gamma mappings. The output of the latter is visibly better than that of the former.}
\label{fig:mapping}
\vspace{-10pt}
\end{figure}

To synthesize an under/over-exposed image, we employ a gamma mapping function, which is a nonlinear operation often used to adjust the overall brightness of an image in the image processing pipeline \cite{gamma_def}. The gamma mapping function is based on the observation that human eyes perceive the relative change in the light following a power-law function rather than a linear function as in cameras \cite{krueger1989reconciling}. The connection between the gamma mapping function and the human visual system enables the gamma mapping function to be widely used in image contrast enhancement~\cite{5370400,rahman2016adaptive,7868607}. However, rather than apply the gamma function directly to the original image, we adopt a haze removal technique in which we apply it to the inverted image to generate $2N$ reference images $Y_n$ as shown in Eq.~\eqref{eq:mapping_function}. The reason is that hazy images and poor lighting images normally share the same property of low dynamic range with the high noise level. Therefore, haze removal techniques (e.g. using an inverse image) can be used to enhance poor lighting images. When negative images are employed, we discovered that the contrast of images may be improved easier, thus producing a more visually pleasant image, as shown in Fig.~\ref{fig:mapping}. In addition, our proposed function also has the same form as the well-known mathematical image processing model LIP~\cite{Jourlin2001LogarithmicIP}, which had been proven with both physical and human visual systems, thus making our mapping function more theoretically sound.
\begin{equation}
    Y_n = 1 - (1 - I)^{\gamma_n},
    \label{eq:mapping_function}
\end{equation}
where $\gamma_n$ is a random number whose logarithm value ${X_n = \log(\gamma_n)}$ is sampled as follows: $X_n \sim U(0, 3)$ for under-exposed reference images and $X_n \sim U(-2, 0)$ for over-exposed reference images.

\subsection{Pseudo Ground-truth Image Generator}
\label{sec:pseudo_gt_generator}
To create the pseudo GT image, we compare and combine the $2N$ generated reference images, the original image, and the output of the enhancement network of the same image in the previous epoch.
The idea behind our approach is inspired by prior work on exposure fusion~\cite{exposure_fusion} where a set of perceptual quality measurements for each pixel are calculated on the reference image sequence. These measurements can encode desired attributes such as brightness, contrast, and saturation and have shown their effectiveness in generating a high-quality high dynamic range (HDR) image from an exposure sequence. Therefore, in this paper, we adopt the high-level concept of these measurements but propose a new formulation for each term and a new way to combine these terms together to produce the pseudo GT image.

\noindent\textbf{Well-exposedness} of a pixel estimates how likely a pixel belongs to a well-exposed region. We use the L1 distance between the average intensity value of a local region to the well-exposed level, which is set to 0.5. Thus, the well-exposedness of a pixel is defined as:

\begin{equation}
    E(x,y) = \left| \mu_{p_{xy}} - 0.5 \right|,
\label{eq:well_exposed}
\end{equation}
where $p_{xy}$ is a patch $K \times K$ centered at $(x,y)$ and  $\mu_{p_{xy}}$ is its mean intensity value. We set $K=25$ in this paper.

\noindent\textbf{Local contrast} is the local variance of all pixels $I(u,v)$ of the patch $p_{x,y}$ 
\begin{equation}
    C(x,y) = \dfrac{1}{K \times K}\sum_{u,v \in p_{x,y}}\left[I(u,v) - \mu_{p_{xy}}\right]^2.
\end{equation}

\noindent\textbf{Color saturation} of a pixel measures its vividness. We use the saturation channel in the HSV color space to measure the color saturation of a pixel. In this color space, the saturation is defined as:
\begin{equation}
    S(x,y) = \dfrac{\max(R, G, B) -  \min(R, G, B)}{\max(R, G, B)},
\end{equation}
where $R, G, B$ correspond to red, green, and blue values of the pixel $(x,y)$.

Since a good-looking output image is one with a low well-exposedness value but high contrast and saturation values, we obtain the final pseudo GT image $T$ by selecting the best regions from all reference images $Y_n$ as follows:
\begin{equation}
\begin{gathered}
    T(x,y) = Y_n(x,y) \\
    \text{with } n = \underset{n=1,...,2N+2}{\mathrm{argmax}}\, \frac{C_n(x,y)S_n(x,y)}{E_n(x,y)}.
\end{gathered}
\end{equation}

It is worth noting that the way we use the final score is completely different from that of \cite{exposure_fusion}. In \cite{exposure_fusion}, the final score map is used as the weight in the weighted sum to combine the reference images together in order to obtain the pseudo GT image. In contrast, we use the final score map as an image comparison tool to select the best regions from the reference images.

\subsection{Image Enhancement Network} \label{enhancement_net}

Similar to \cite{zerodce}, our network will learn to predict the intermediate parameters of a tone mapping function instead of directly predicting the output image. We design our lightweight enhancement network based on the UNet architecture~\cite{ronneberger2015unet}. 
 
The detail of network architecture is provided in the \textit{Supp. material}. For consistency with the reference image generation module, the invert gamma mapping function in Sec.~\ref{refence_generator} is used to produce the final image $Y$ given the predicted gamma map $\gamma$ and the original image I: 
\begin{equation}
    Y = 1 - (1-I)^{\gamma}.
\end{equation}

We then train our model end-to-end to minimize the following loss function:
\begin{equation}
L = L_{rec} + \alpha L_{tv},
\end{equation}
where $L_{rec}$ and $L_{tv}$ are the reconstruction loss and total variation loss, respectively, and $\alpha$ is a coefficient to balance between the two losses.

\noindent\textbf{Reconstruction loss.} We adopt the mean squared error between the network prediction and the pseudo GT image as follows:
\begin{equation}
    L_{rec} = \dfrac{1}{3HW} \sum_{c,x,y}\left[\hat{Y}(c,x,y) - T(c,x,y)\right] ^ 2,
\end{equation}
where $c$ is the color channel, $\hat{Y}$ is the output image and $T$ is our generated pseudo GT image in Sec.~\ref{sec:pseudo_gt_generator}; $H$ and $W$ are the height and width of the input image, respectively.

\noindent\textbf{Total variation loss.} In the homogeneous areas, the adjacent gamma values should be similar to avoid sudden changes, which can create visual artifacts. Therefore, we apply a familiar smoothness prior to image restoration tasks, called total variation minimization~\cite{Rudin1992NonlinearTV,Ng2011ATV}, to the predicted gamma map. The total variation loss is defined in Eq.~\eqref{eq:tv_loss} 

\begin{align}
    \mathcal{L}_{tv} = \frac{1}{3HW}\sum_{c,x,y} & \{ |\gamma_c(x+1,y) - \gamma_c(x,y)| + \nonumber \\ 
    &|\gamma_c(x,y+1)- \gamma_c(x,y)| \},
    \label{eq:tv_loss}
\end{align}
where $\gamma_c$ is the predicted gamma value corresponding to the color channel $c$.

\section{Experiments} \label{sec_experiments}


\noindent\textbf{Datasets.} 
We assess our approach as well as comparative methods on two main multi-exposure datasets: Afifi (introduced by Afifi \etal{}~\cite{afifi2021learning}) and SICE~\cite{sice} datasets. The Afifi dataset contains 24,330 sRGB images rendered from the MIT-Adobe FiveK dataset \cite{fivek} by varying their digital exposure settings. The SICE dataset has two parts 1 and 2 with 360 and 229 multi-exposure sequences (sets of images of the same scene captured at different exposure levels), respectively. We employ part 1 as the training set and part 2 as the test set. For generalization evaluation, we also test all approaches on the LOL dataset~\cite{retinexnet}, which is composed of 500 pairs of low-light and normal images.

\noindent\textbf{Implementation details.} We train our image enhancement network on an NVIDIA A100 GPU, using the Adam optimizer with a batch size of 64. In the SICE dataset, our model is trained with 140 epochs while the number of epochs for training the Afifi dataset is 30. The learning rate is $5e^{-4}$ and is reduced by half on a plateau with the patience of 5. All the input images are resized to $256\times 256$ during training. 
The coefficient of the total variation loss $\alpha$ for training on each dataset and other implementation details are empirically selected and reported in the \textit{Supp. material}.

\subsection{Comparison with Prior Work}
We compare our approach with two traditional methods: CLAHE \cite{clahe}, IAGCWD \cite{IAGCWD}, one unpaired method EnlightenGAN \cite{enlightengan}, and two unsupervised methods: ZeroDCE~\cite{zerodce}, Zheng and Gupta~\cite{semanticguided}. We also include the two supervised methods: HDRNet~\cite{hdrnet}, Afifi \etal{}~\cite{afifi2021learning} for the reference purpose only. The results of these methods are reproduced by using their public source codes with the recommended parameters.

\begin{table*}[t]


\centering
\small
\begin{tabular}{l|cc|cc|cc|cc|cc}
\hline
\textbf{Method} & \multicolumn{6}{c|}{\textbf{Afifi}} & \multicolumn{2}{c|}{{\textbf{SICE}}} & \multicolumn{2}{c}{{\textbf{LOL}}}  \\
 \cline{2-7}
 & \multicolumn{2}{c|}{Under} & \multicolumn{2}{c|}{Over} & \multicolumn{2}{c|}{Full} & \multicolumn{2}{c|}{} \\
 \cline{2-11}
 & PSNR & SSIM & PSNR & SSIM & PSNR & SSIM & PSNR & SSIM & PSNR & SSIM \\
\hline
\rowcolor{gray}
HDRNet (S) & 19.35 & 0.817 & 20.65 & 0.846 & 20.13 & 0.834  &17.25  &0.683  & 17.29 & 0.766  \\
\rowcolor{gray}
Afifi \etal{} (S) & 18.88 & 0.845 & 19.05 & 0.850 & 18.98 & 0.848  & - & - & - & -\\
\hline
CLAHE  (N)  & 16.67 & 0.780 & {18.19} & {0.806} & {17.58} & {0.796} & 13.89 & 0.610 & 10.02 & 0.427\\
IAGCWD  (N)  & 13.23 & 0.681 & 17.99 & 0.820 & 16.08 & 0.765  & 14.09 & 0.635 & 11.26 & 0.519 \\
\hline
ZeroDCE (U) & 15.36 & 0.783 & 11.85 & 0.739 & 13.25 & 0.757  & 14.28 & 0.657 & 14.16 & 0.654\\
Zheng and Gupta (U)  & {16.69}  & {0.806} & 11.58 & 0.726 & 13.62 & 0.758 & 12.54 & 0.626 & {14.89} & {0.675}\\
EnlightenGAN (U*) & 14.28 & 0.752 & 14.05 & 0.766 & 14.14 & 0.762  &	{14.60} &	{0.680} & 12.32 & 0.596 \\
\hline
\Approach~(U)  & \textbf{18.82} & \textbf{0.858} &\textbf{19.72} & \textbf{0.875} &\textbf{19.36} & \textbf{0.869} & \textbf{17.74} & \textbf{0.704} & \textbf{16.60} & \textbf{0.693} \\
\end{tabular}
\vspace{5pt}
\caption{Results on SICE, Afifi, and LOL dataset. The higher the better. The best results are in bold. The terms ``under" and ``over" stand for under-exposure and over-exposure subsets. The terms ``N", ``U", ``U*" ``S" stand for non-learning, unsupervised, unpaired, and supervised, respectively. All methods are trained with the same training sets but different supervision levels. Due to the Matlab license issue, we could not train and evaluate the performance of Afifi \etal{} on the SICE and the LOL datasets. Note that HDRNet~\cite{hdrnet} and Afifi et al.'s method~\cite{afifi2021learning} are both supervised methods (faded in gray, solely for reference purpose).}
\label{table:quantity_result}
\end{table*}

\noindent\textbf{Objective image quality assessment.} We adopt two standard referenced metrics: peak signal-to-noise ratio (PSNR) and structural similarity index measure (SSIM) in the cases where the ground-truth images are available in the testing set (for evaluation purposes only). To see the effectiveness of our method on each type of lighting condition, we also report the metric scores on lighting-dependent subsets of the Afifi dataset \cite{afifi2021learning}, which includes 3,543 over-exposed images and 2,362 under-exposed images. 

Table \ref{table:quantity_result} reports the quantitative results obtained by each method on the Afifi, SICE, and LOL datasets. On the Afifi dataset, our approach outperforms all state-of-the-art unpaired model EnlightenGAN~\cite{enlightengan} and unsupervised models ZeroDCE~\cite{zerodce}, and ZeroDCE++~\cite{dce_extension} with significant margins (+5 and +0.1 in PSNR and SSIM metrics, respectively). 
On the SICE dataset, our approach also surpasses all other unsupervised methods on the SICE dataset with large margins (+3 and +0.02 in PSNR and SSIM metrics, respectively). When compared with the supervised methods, surprisingly, the proposed method obtains better results than HDRNet~\cite{hdrnet} and Afifi \etal{}~\cite{afifi2021learning} in the SSIM index. We further assess the generalization abilities of all methods on the LOL dataset. In this experiment, we report the results of all methods trained on the SICE dataset without further tuning. As can be seen in Tab.~\ref{table:quantity_result}, the same trend can be observed in which we outperform all unsupervised and unpaired approaches and perform slightly worse than a supervised model HDRNet~\cite{hdrnet}. However, HDRNet~\cite{hdrnet} tends to produce output images with visual artifacts that are shown in the next section.

\begin{figure*}[t]
\centering
\includegraphics[width=0.9\textwidth, page=3]{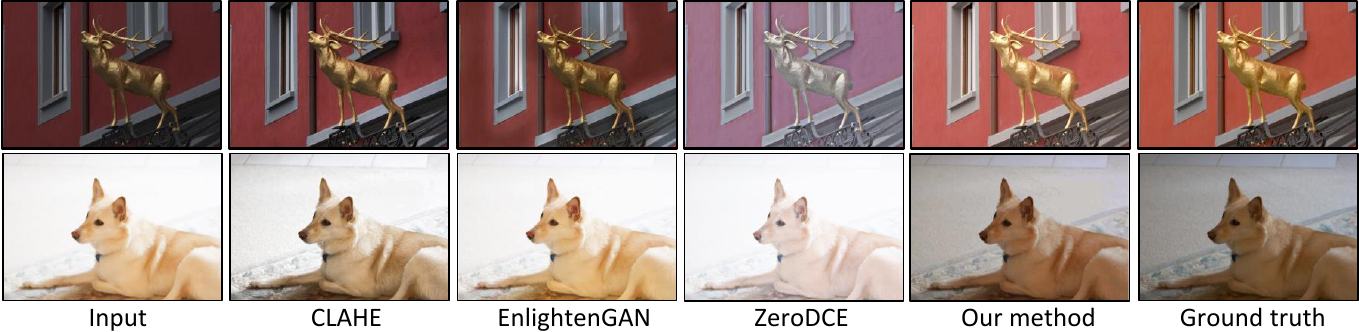}
\caption{Visual comparison with unsupervised and traditional methods. 
In the under-exposure situation, CLAHE~\cite{clahe} and EnlightenGAN~\cite{enlightengan} could not brighten the image to a proper level, while ZeroDCE~\cite{zerodce} tends to produce an image with washed-out colors. As for the over-exposed image, only CLAHE seems to produce an acceptable output whereas  ZeroDCE and EnlightenGAN appear to fail to recover the image’s details in such a condition.
}
\label{fig:visual_unsupervised}
\end{figure*}

\begin{figure}[t]
\centering
\includegraphics[width=\columnwidth, page=4 ]{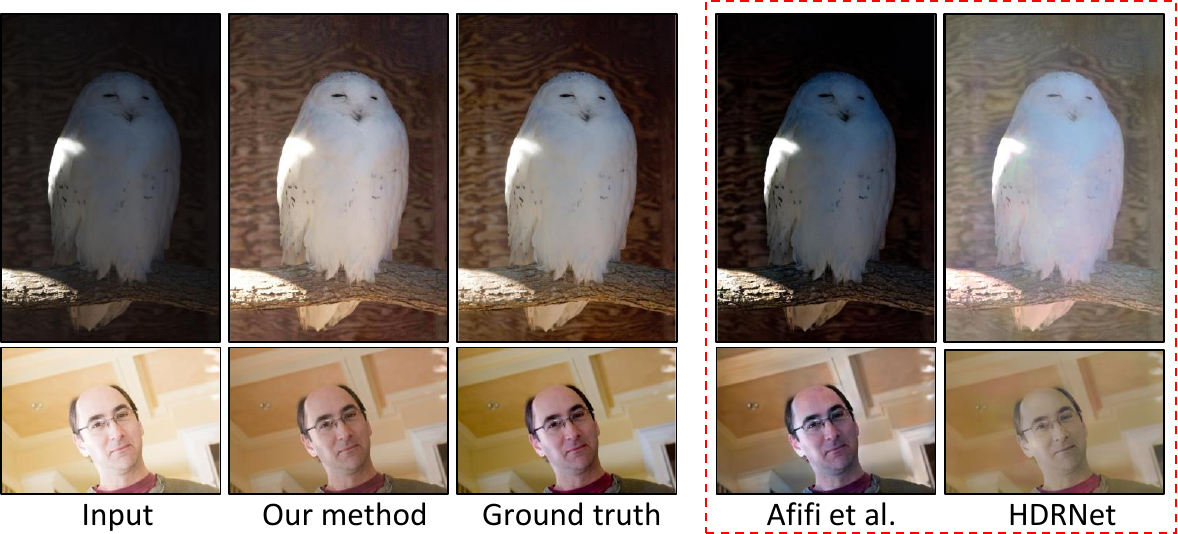}
\caption{Visual comparison with supervised methods (put in red box). 
Although HDRNet~\cite{hdrnet} achieved the best PSNR and SSIM scores in most of the cases, it often produces output images with visible artifacts.}
\label{fig:visual_supervised}
\vspace{-5pt}
\end{figure}

\noindent\textbf{Subjective image quality assessment.} 
A visual comparison among unsupervised approaches with typical under-exposure and over-exposure scenes is presented in Fig.~\ref{fig:visual_unsupervised}. Our model is the only one that works on both under- and over-exposure. In the case of underexposure, CLAHE~\cite{clahe} and EnlightenGAN~\cite{enlightengan} could not brighten the image to a proper level, while ZeroDCE~\cite{zerodce} tends to produce an image with washed-out colors. Regarding the over-exposure situation, only CLAHE seems to produce a decent output image whereas ZeroDCE and EnlightenGAN appear to fail to recover the image’s details in such a condition.
A visual comparison with other supervised methods is also shown in Fig.~\ref{fig:visual_supervised}. As mentioned previously, although HDRNet~\cite{hdrnet} achieved the best PSNR and SSIM scores in most of the cases, however, it often produces output images with visible artifacts (shown in Fig.~\ref{fig:visual_supervised}).

\begin{figure}[t]
\centering
\includegraphics[width=0.41\textwidth, page=5]{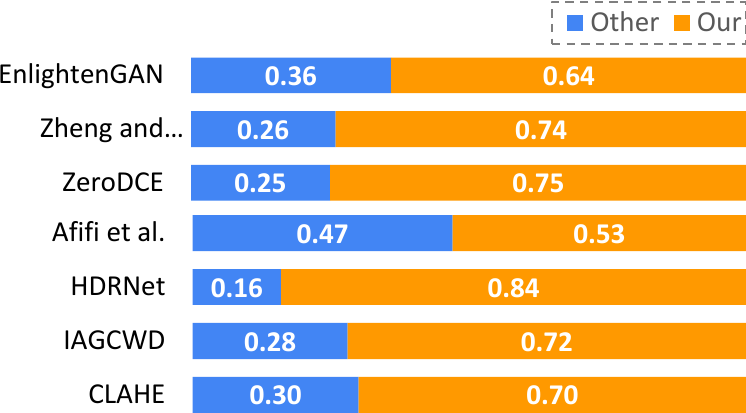}
\caption{Results of user study - Others vs Ours. The blue color shows the preference percentages of the other methods, while the orange color shows ours.}
\label{fig:user_study}
\end{figure}

\noindent\textbf{User study.} For a more convincing evaluation, we also conduct a user study with 260 participants on 100 scenes from the testing set to assess human preference for the enhanced results. Out of 100 scenes, 30 scenes are randomly selected to show for each participant, and for each scene, our enhanced image along with another image, which is a result of a random method, is presented. We believe that showing the results of two methods at a time is more reliable than showing the results of all eight methods and asking the users to either rank all methods or choose the best one only. The former is error-prone since the users need to rank pairwise $C(8, 2)=28$ times for each question. The latter is not informative if ours is not the best. We restrict the sampling method to ensure that all the methods appear evenly across user responses.  The participants then are asked to pick a better image in each pair based on the three following criteria: (1) whether all parts in the image are clearly visible; (2) whether the result introduces any color deviation; and (3) the better image based on their preference. The detailed comparison between ours and other methods is reported in Fig.~\ref{fig:user_study}. As can be seen, our enhanced images are preferred in all cases including both supervised and unsupervised approaches, only Afifi et al.'s approach~\cite{afifi2021learning}  has a preferable ratio that is relatively comparable to ours. HDRNet~\cite{hdrnet} is less desirable in our user study due to their unpleasant output images.

\subsection{Ablation Study} 

In this section, we conduct experiments on analyzing the stability of our method and the impact of different components of our proposed framework. 
Other experiments related to hyper-parameter selection are presented in the \textit{Supp. material.}

\begin{figure}[t]
\small
\centering
\includegraphics[width=.35\textwidth, page=1]{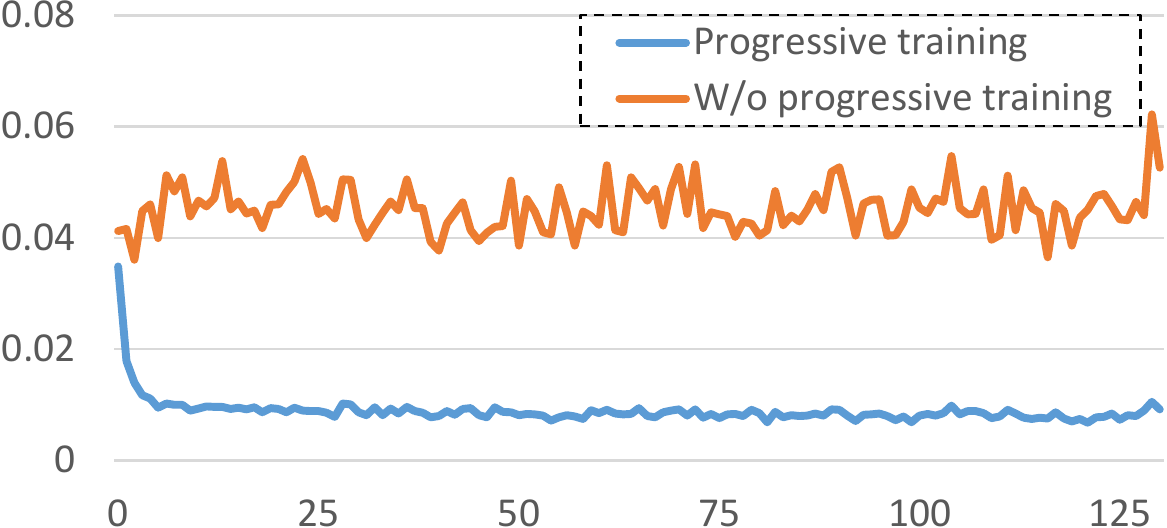}
\caption{Mean squared error (MSE) between pseudo GT images at two consecutive epochs with and without PTS.}
\label{fig:training_stability}
\end{figure}

\begin{figure}[t]
\small
\centering
\includegraphics[width=\columnwidth, page=10]{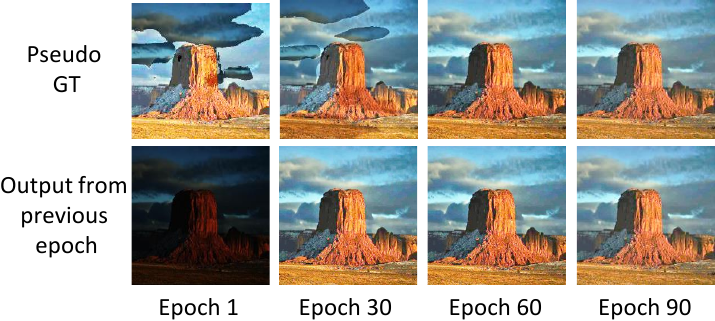}
\caption{Examples of the pseudo GT image and model output from previous epoch while training. Note that, for epoch 1, the output in the previous epoch is equivalent to the input image. For the first few epochs, the pseudo GT is unstable due to its dependence on the randomness of the reference image generator. However, as the quality of the model's output increases and surpasses reference images, the pseudo GT image tends to converge thanks to the PTS. In addition, the output of our model in these epochs is very close to the pseudo-GTs except for some hash regions, providing useful attention for our model at the latter training stage.}
\label{fig:visual_training_stability}
\end{figure}

\noindent\textbf{Training stability.} 
Since our method relies on a random reference image generator to produce pseudo GT images, a concern might be raised that whether or not this random factor affects our model's performance. To answer this question, we have retrained our model 10 times with different random seeds. The average performance on SICE is $17.69 \pm 0.11$ in PSNR and $0.704 \pm 0.0013$ in SSIM showing that our method is stable regardless of random sampling. This stability is achieved through the progressive training strategy (PTS). In Fig.~\ref{fig:training_stability}, with the PTS, the MSE between two consecutive pseudo GT decreases when the number of training epochs increases, suggesting that the PTS ensures training convergence and stability. This assumption are also demonstrated in Fig.~\ref{fig:visual_training_stability}. As a result, our model's final performance also gets better with approximately $+1.5$ and $+0.02$ in terms of PNSR and SSIM, respectively in comparison with the normal training method.

\noindent\textbf{The influence of pseudo GT generator in other enhancement networks.}
We analyze the influence of our pseudo GT generator by replacing our Enhancement Network with the network presented in ZeroDCE~\cite{zerodce} and  EnlightenGAN~\cite{enlightengan}, and retrain these models on the SICE dataset. As demonstrated in Tab.~\ref{table:pseudo_GT}, using our proposed generator 
significantly improves the overall performance of these two networks on the SICE dataset (+1 in PSNR). Qualitative results are provided in the \textit{Supp. material}.

\begin{table}[t]
    \centering
    \small
   \begin{tabular}{lcccc} 
    \hline
    \textbf{Method} & \textbf{PSNR} \\ 
    \hline
    ZeroDCE & 14.28 \\
    ZeroDCE + our pseudo GT & 15.30 \\
    \hline
    EnlightenGAN & 14.60 \\
    EnlightenGAN + our pseudo GT & 15.34 \\
    \end{tabular}
    \vspace{5pt}
    \caption{The influence of our pseudo GT generator on ZeroDCE~\cite{zerodce} and EnlightenGAN~\cite{enlightengan} on the SICE dataset. Our suggested approach also improves these two networks' abilities in handling over-exposure cases, resulting in a considerable improvement (+1 in PSNR).}
    \label{table:pseudo_GT}
\end{table}

\begin{figure}
\centering
\includegraphics[width=0.4\textwidth,page=1]{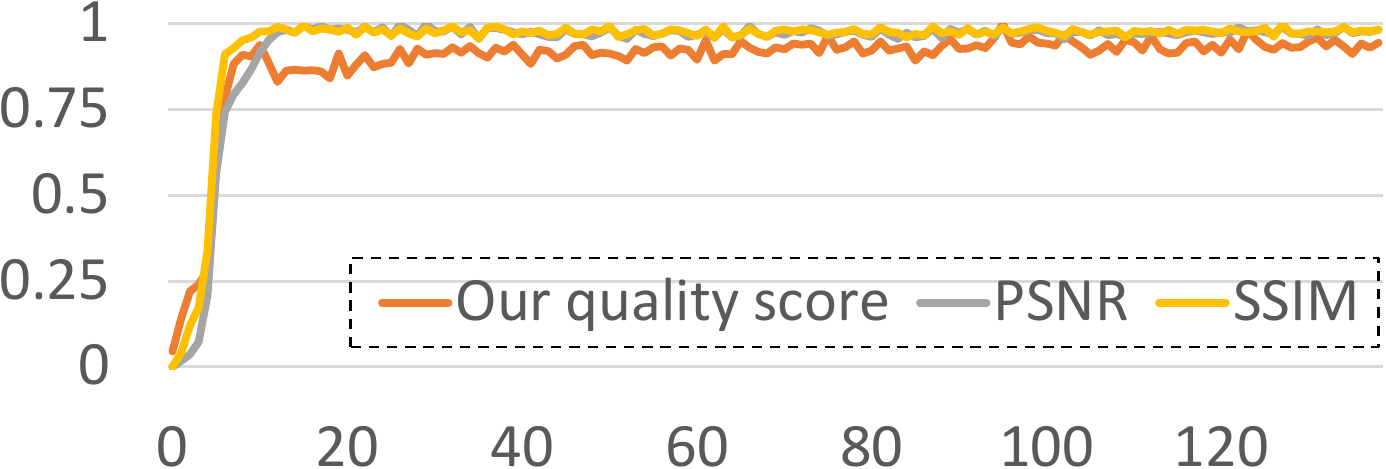}
\caption{Average PSNR, SSIM (compared with provided GT images), and our proposed image quality score of our pseudo-GT over training epochs (x-axis). 
}
\vspace{-5pt}
\label{fig:IQA}
\end{figure}

\begin{figure}[t]
    \centering
    \includegraphics[width=.45\textwidth, page=7]{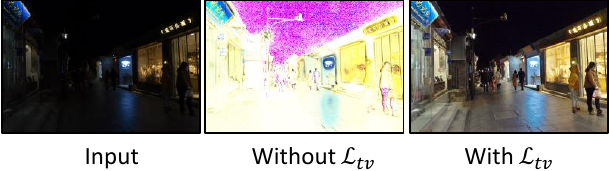}
    
    \caption{The influence of the total variation loss. Without the loss, the gamma values of neighboring regions are not so smooth, thus breaking down the image structure.}
    \label{ablation_tvloss}
\end{figure}
\noindent\textbf{The correlation between our proposed quality score and the similarity between the pseudo GT images and reference GT images.} We conduct additional experiments to evaluate the correlation between our proposed quality score and the similarity between pseudo GT images and reference GT images measured by PSNR and SSIM metrics. The results in Fig. \ref{fig:IQA} show that our quality score is an effective measurement of image quality without GT images.

\noindent\textbf{Contribution of total variation loss.} We also present the results of our enhancement network trained with the absence of total variation loss in Fig.~\ref{ablation_tvloss}. Without this loss, our model tends to break the relation between neighboring regions, thus breaking down the image structure.

\subsection{Computational cost comparison}
\begin{table}[]
\centering
\small
\begin{tabular}{lccc} 
\hline
\textbf{Method} & \textbf{RT (ms)} & \textbf{\#Params } & \textbf{\#GMACs}\\ 
\hline
EnlightenGAN & 94.38 & 8,636,675 & 197.11\\ 
ZeroDCE & 37.87 & 79,416 & 61.59 \\ 
Zheng and Gupta & 36.86 &10,561 &7.834  \\ 
\hline
Our method & 20.08 & 15,251 & 1.804  \\
\end{tabular}
\caption{Running time (\textbf{RT}), \textbf{\#} of parameters (\textbf{\#Params}), and \textbf{\#} of multiply–accumulated operations (\textbf{\#GMACs}). 
}
\label{tab:runtime}
\end{table}

We evaluate the computational cost of our model and other methods  and report the results in Tab. \ref{tab:runtime}. The runtime is measured on on Tesla T4 GPU by processing 50 images of size $1080 \times 720$. The number of multiply–accumulated operations (MACs) and the number of trainable parameters for each network are also presented. As we can see, our method is the fastest and extremely lightweight, making it very suitable for real-time applications.

\section{Application} \label{supp_application}

In this section, we conduct experiments to evaluate the usefulness of our approach on the face detection task for both under- and over-exposure cases. To the best of our knowledge, there is no public face dataset that contains sufficient samples from both under- and over-exposure for validation. Therefore, we synthetically create a new face dataset from the FDDB dataset \cite{fddb} by generating new images with different gamma values. The Dual Shot Face Detector (DSFD) \cite{dsfd} trained on the WIDER FACE dataset \cite{yang2016wider} is used as a pre-trained face detector. More concretely, we feed the images enhanced by several different image enhancement methods to the pre-trained face detector and observe its performance changes.

\begin{figure}[]
    \centering
    \includegraphics[width=0.5\textwidth, page=8]{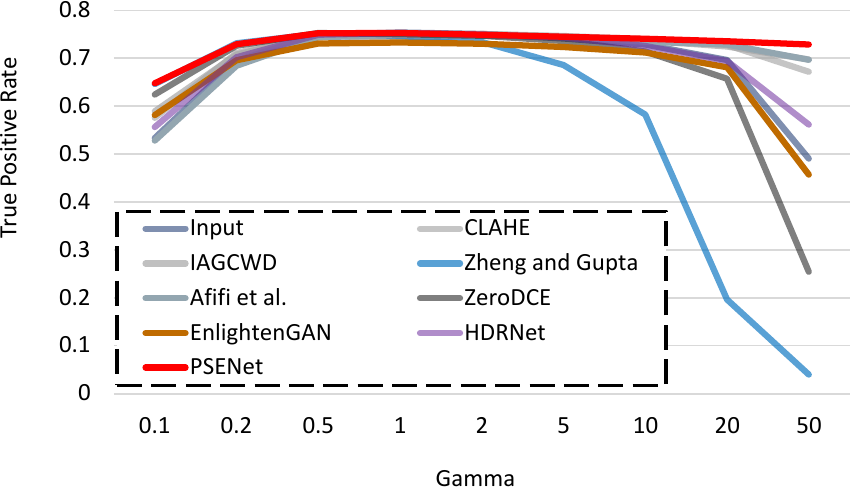} 
    \vspace{1pt}
    \caption{True positive rate when \# of false-positive samples equals 500 for different gamma values on the FDDB dataset \cite{fddb}.}
    \label{fig:FDDB_TPR}
    
\end{figure}

Fig.~\ref{fig:FDDB_TPR} depicts the true-positive rate when the number of false-positive samples equals 500, which is computed by the evaluation tool provided in the FDDB dataset~\cite{fddb}. 
As can be seen, with our image enhancer, DSFD~\cite{dsfd} achieves better metric scores consistently on both too dark (low gamma value) and too bright (high gamma value) images. Meanwhile, other methods such as ZeroDCE~\cite{zerodce}, Zheng and Gupta \cite{semanticguided} and EnlightenGAN ~\cite{enlightengan} perform poorly in the over-exposure cases, resulting in a decrease in face detection performance. This demonstrates the robustness of our method under various lighting conditions.

 \begin{figure}[]
    \centering
    \includegraphics[width=.48\textwidth, page=9]{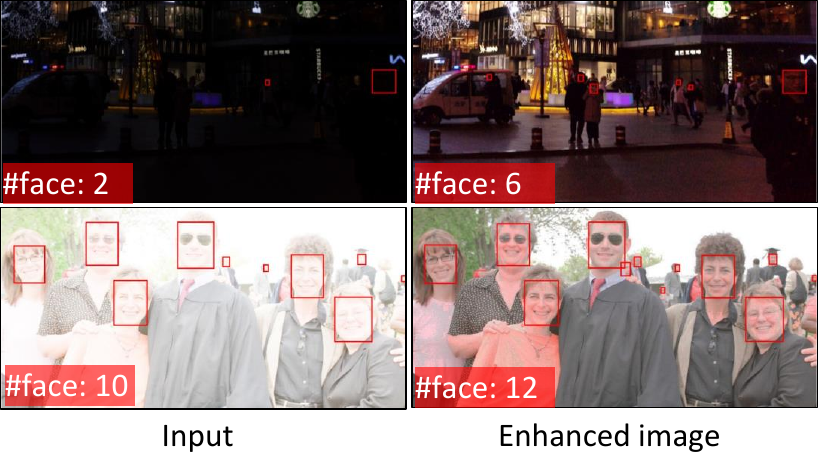}
    \caption{Outputs of the detection model on the images before and after preprocessed by our method. \#face denotes the number of faces that are successfully detected by the DFSD. }
    \label{fig:pr_curve}
\end{figure}

We also present the output of DFSD on two real images where our model is utilized as a preprocessing module in Fig.~\ref{fig:pr_curve}. As can be seen, our model can recover the image detail in both extremely dark or over-bright regions, thus improving the performance of the face detector.

\section{Conclusion}

We have introduced a novel progressive self-enhancement network \Approach~for image enhancement that is robust to a variety of severe lighting conditions, including under-exposure and over-exposure. In particular, we have developed a new method for generating effective pseudo GT images for training our extreme-light enhancement network in an unsupervised manner. As a result, the burden of gathering the matched photographs is removed.
Our extensive experiments show that the proposed approach consistently outperforms previous unsupervised methods by large margins on several public datasets and obtains comparable results with supervised counterparts.
We also demonstrate the superior performance of \Approach~over all other approaches in the application of face detection in both under-exposure and over-exposure settings.
These results justify the importance of \Approach~not only in pleasing human vision but also in improving machine perception.


\thispagestyle{empty}

\section{Supplementary Materials}

In this supplementary material, we provide implementation details of our proposed method and additional results  which are not included in the main paper due to the space limitation. 

\subsection{Implementation Details} \label{supp_implementaion}
\begin{figure}[ht!]
\centering
\includegraphics[width=0.5\textwidth, page=7]{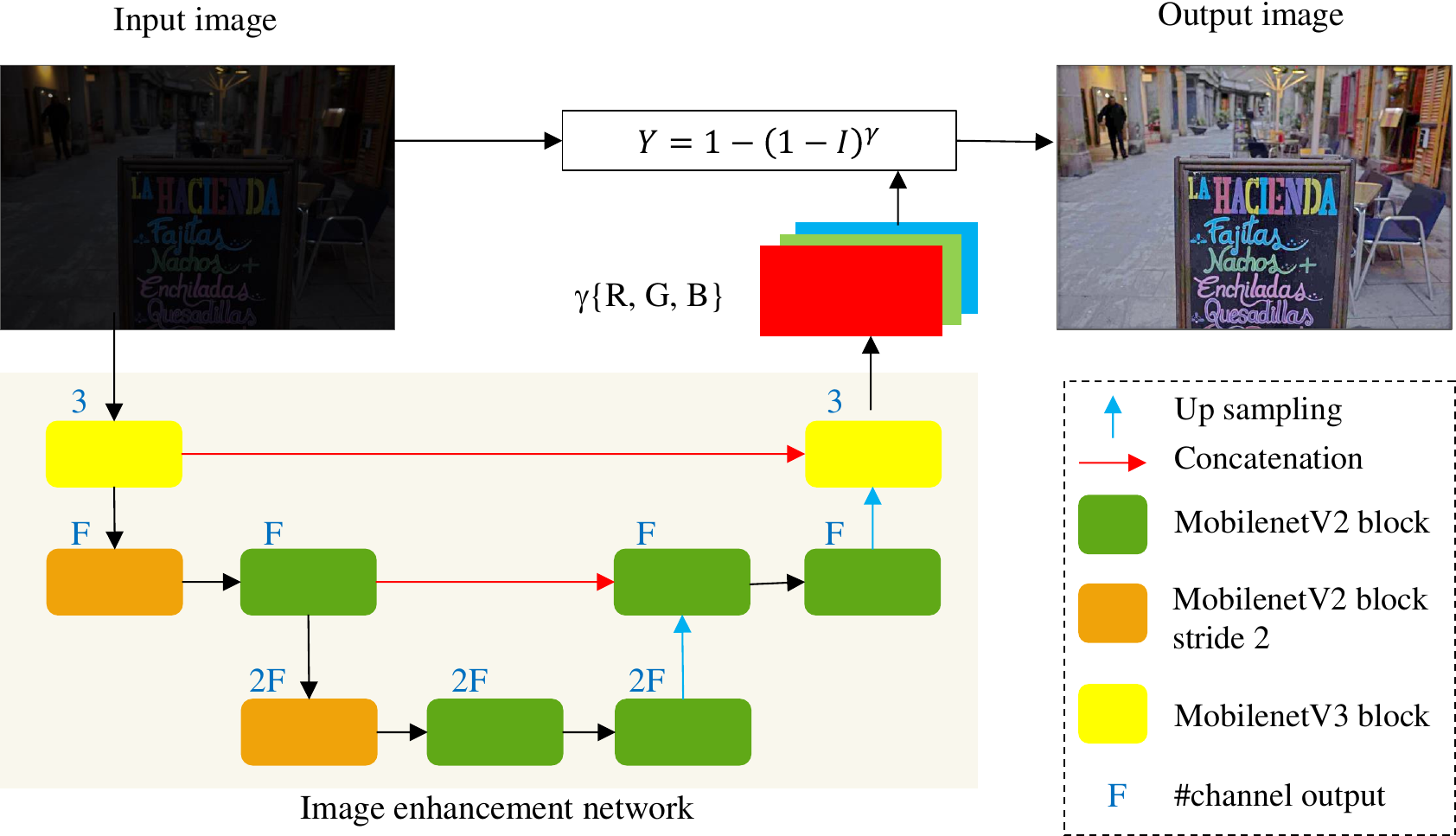}
\caption{Image enhancement network in detail. A lightweight UNet architecture is employed to predict the gamma map $\gamma$ for each channel. The enhanced image is obtained by applying the gamma mapping function with the predicted gamma map}
\label{fig:enhancement_network}
\vspace{-5pt}
\end{figure}

\noindent\textbf{Image Enhancement Network.} As described in the main paper, we employ a lightweight UNet architecture~\cite{ronneberger2015unet} as illustrated in Figure~\ref{fig:enhancement_network} to build up our network. The specification of our network is given in Table~\ref{tab:architecture_specification}.
\begin{table}[h]
\setlength{\tabcolsep}{3pt}
    \caption{Architecture detail of the image enhancement network. \#Output denotes the number of output channels.}
    \centering
    \begin{tabular}{ccccc}
    \hline
        Input & Expand size & \#Output & MobileNet & Stride  \\
        \hline
        $256^2 \times 3$ & 6 & 3 & V3 & 1 \\
        $256^2 \times 3$ & 24 & 16 & V2 & 2 \\
        $128^2 \times 16$ & 24 & 16 & V2 & 1 \\
        $128^2 \times 16$ & 48 & 32 & V2 & 2 \\
        $64^2 \times 32$ & 48 & 32 & V2 & 1 \\
        $64^2 \times 32$ & 48 & 16 & V2 & 1 \\
        $128^2 \times 32$ & 48 & 16 & V2 & 1 \\
        $128^2 \times 16$ & 24 & 3 & V2 & 1 \\
        $256^2 \times 6$ & 9 & 3 & V3 & 1 \\
    \end{tabular}
    
    \label{tab:architecture_specification}
\end{table}

\noindent\textbf{Training Process.} Our proposed approach is implemented using PyTorch framework. We train our image enhancement network on an NVIDIA A100 GPU from scratch, using the Adam optimizer with a batch size of 64. The learning rate is $0.0005$ and is reduced by half on plateau with the patience of 5. The input images are resized to $256\times 256$ without applying any augmentation techniques. For the SICE dataset, our model is trained for 140 epochs with the coefficient of the total variation loss $\alpha$ being 5. For the Afifi dataset, the number of training epochs is 30 and $\alpha$ is set to 500.

\begin{figure}[]
\centering
\includegraphics[width=.47\textwidth, page=8]{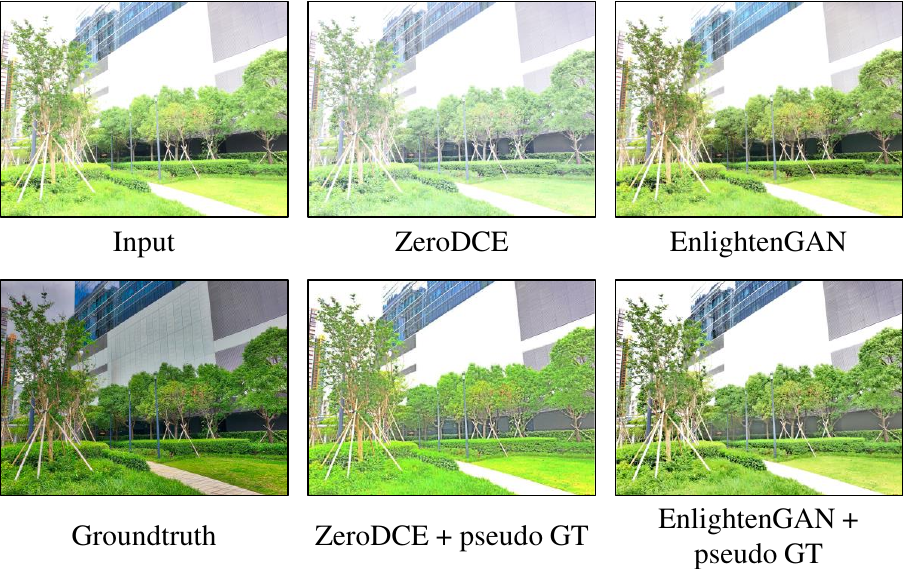}
\caption{The influence of the pseudo GT generator on ZeroDCE~\cite{zerodce} and EnlightenGAN~\cite{enlightengan}. Our proposed approach also improves these two networks' abilities of handling over-exposure cases}
\label{fig:pseudo_GT}
\end{figure}
\subsection{Ablation Study} \label{supp_ablation}

\noindent\textbf{The influence of pseudo GT image generator.}
As stated in the main paper, our training strategy also shows its effectiveness when combined with other image enhancement networks. Specifically, we apply our training strategy to the network architecture of ZeroDCE~\cite{zerodce} and EnlightenGAN~\cite{enlightengan} with other settings kept unchanged. The results shown in Figure~\ref{fig:pseudo_GT} demonstrate that our training strategy is robust to the network architecture selection when consistently improving the performance.

\noindent\textbf{The impact of the number of random reference images.}
We further evaluate our model's performance when adjusting the number of random reference images. The results are presented in Table~\ref{table:Random_reference}. We empirically find that increasing the number of random references improves the quality of the output images in the SICE dataset. However, with the Afifi dataset, it might have a negative impact on network performance. Thus, this hyper-parameter is dataset-specific. 
\begin{table}[t]
    \centering
    \begin{tabular}{lcccc} 
    \hline
    {{\textbf{Method}}} & \multicolumn{2}{c}{\textbf{SICE}} & \multicolumn{2}{c}{\textbf{Afifi \etal{}}}\\ 
    \cline{2-5}
    {} & PSNR & SSIM  & PSNR  & SSIM\\
    \hline
    N = 1 & 17.74 & 0.704 & 19.36 & 0.869 \\
    N = 3 &  17.76 & 0.702 & 19.15 & 0.865 \\
    N = 5 &  17.84 & 0.706 & 18.61 & 0.856 \\
    \end{tabular}
    \caption{The impact of the number of randomly generated reference N to the final performance of our approach on SICE~\cite{sice} and Afifi~\cite{afifi2021learning} datasets.}
    \label{table:Random_reference}
\end{table}

\begin{figure}[t]
\centering
\includegraphics[width=.45\textwidth, page=6]{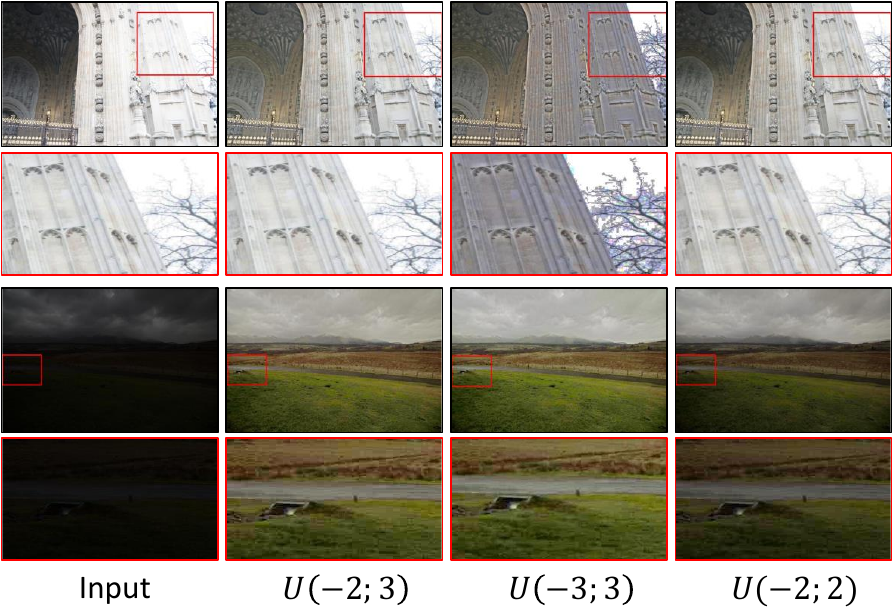}
\label{fig:Random_range}
\caption{The impact of the range for sampling random reference images. $U(a;b)$ indicates the range to sample brightness factor $X$. The first and third rows show the whole images, while the second and last rows show the corresponding close-ups.}
\end{figure}

\noindent\textbf{Impact of the range for sampling reference images.} 
In terms of brightness modification, we found that the best range to sample the reference images is from 0 to 3 for darker image generation and from -2 to 0 for synthesizing brighter images. 
If we narrow the range for under-exposure to $(0,2)$, our model's performance decrease noticeably. The reason is that the produced gamma map is then limited, thus, our model could not increase the brightness of the input image to a proper value in extreme cases, as demonstrated in the two last rows of Fig.~\ref{fig:Random_range}. 
On the other hand, regarding the range of sampling brighter images, extending this range from $(-2, 0)$ to $(-3, 0)$ might create undesired artifacts in overexposed areas. Due to image clipping, the color information in these areas is not well preserved. Therefore, when reducing image brightness, instead of producing a vivid image, our model tends to modify the color tone of the input image to gray, which is visually unpleasant.

\begin{figure*}[]
\centering
\includegraphics[width=\textwidth,page=9]{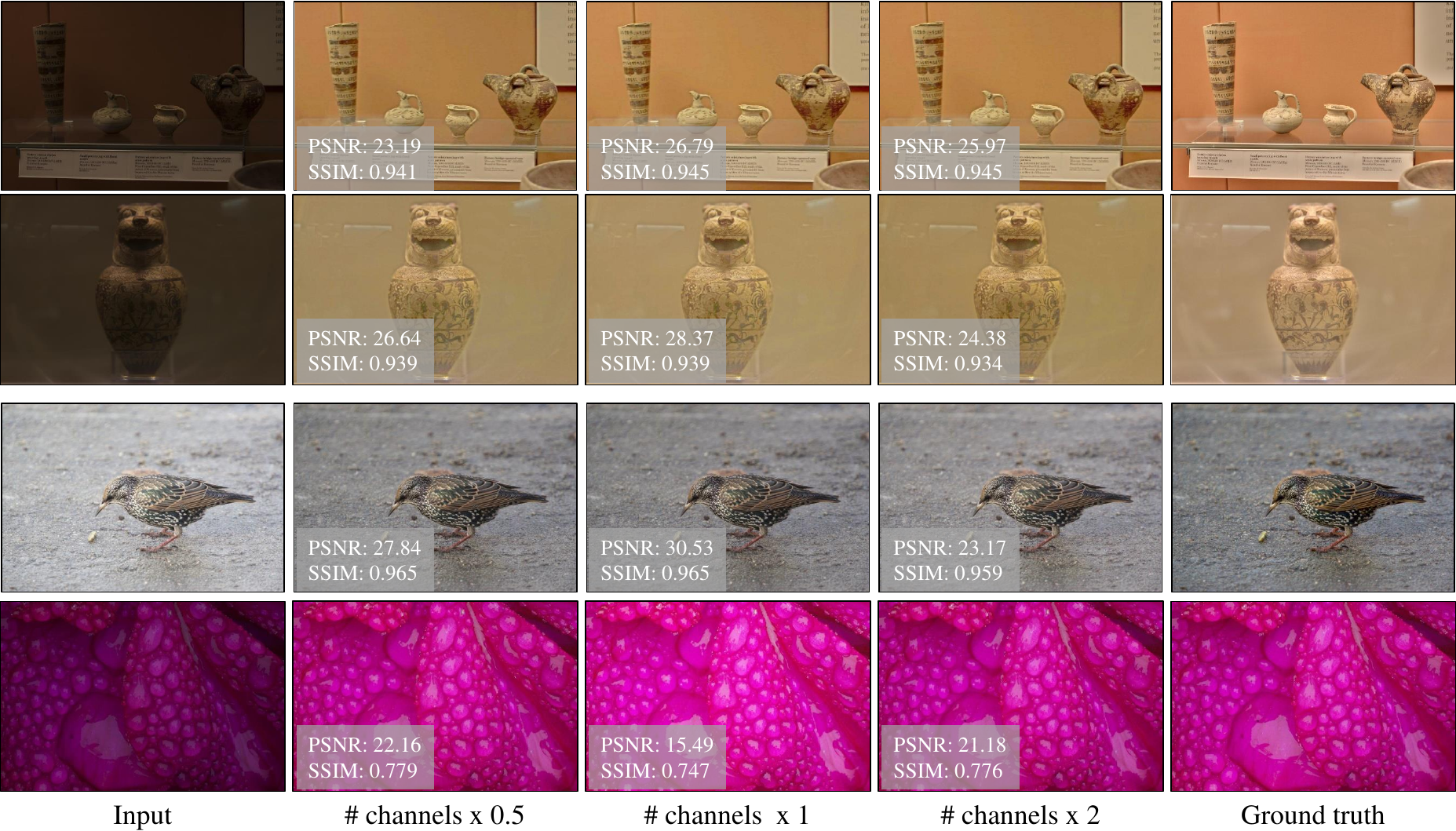}
\caption{Samples whose PSNR values varied most when the number of network parameters is changed. \# channels represents the number of feature maps in each layer of the proposed network (except the first and last layers). It seems that the change in PSNR value is mostly caused by the shift in the brightness level of the output images compared to the ground truths}
\label{fig:network_size}
\end{figure*}

\noindent\textbf{The impact of the network size.} We examine how our image enhancement network performs when the number of trainable parameters is increased or decreased. The quantitative results are shown in Table~\ref{tab:network_size} and qualitative examples are visualized in Figure~\ref{fig:network_size}. Although the quantitative results vary slightly, we do not observe any obvious failure cases when visually comparing the output images. The difference in quantitative results appears to be caused by the shift in the brightness level of the output images compared to the ground truths. However, such output images are still acceptable when analyzed by humans. 
\begin{table}[]
    \centering
    \begin{tabular}{lcccc} 
        \hline
        {{\textbf{Method}}} &  \multicolumn{2}{c}{\textbf{SICE}} & \multicolumn{2}{c}{\textbf{Afifi \etal{}}}\\ 
        \cline{2-5}
        {} & PSNR & SSIM  & PSNR  & SSIM \\
        \hline
        \# channels $\times$ 0.5 & 17.58  & 0.703 & 19.36 & 0.869\\
        \# channels $\times$ 1 & 17.74 & 0.704 & 19.36 & 0.869\\
        \# channels $\times$ 2 & 17.59 & 0.702 & 19.29 & 0.868\\
        \end{tabular}
    \caption{The performance in PSNR and SSIM with different network parameters. The higher the better. \# channels represents the number of channels in each layer of the proposed network (except the first and last layers)}
    \label{tab:network_size}
\end{table}
\begin{table}[]
    \centering
    \begin{tabular}{lcccc} 
        \hline
        {{\textbf{Method}}} &  \multicolumn{2}{c}{\textbf{SICE}} & \multicolumn{2}{c}{\textbf{Afifi \etal{}}}\\ 
        \cline{2-5}
        {} & PSNR & SSIM  & PSNR  & SSIM \\
        \hline
        $\mu = 0.4$ & 16.02  & 0.690 & 17.97 & 0.844\\
        $\mu = 0.5$ & 17.74 & 0.704 &  19.36 & 0.869\\
        $\mu = 0.6$ & 16.60 & 0.6923 & 18.37 & 0.855\\
        \end{tabular}
    \caption{The performance in PSNR and SSIM with different well-exposed levels $\mu$ . The higher the better}
    \label{tab:exposure_level}
\end{table}

\noindent\textbf{Comparison with an image fusion method}. Although our pseudo GT generator's design are inspired by the high level idea of the work introduced in \cite{exposure_fusion}, there are some noticeable differences between ours and theirs including our new quality score and our image combination strategy. We present the comparison between our method and the method in \cite{exposure_fusion} when they are used inside pseudo GT generator module in Table~\ref{ablation_fusion_method}. The results suggest that our proposed design are more effective than the prior work.

\begin{table}[t]
    \centering
        \small
    \begin{tabular}{lcccc} 
    \hline
    \textbf{Method} & \textbf{PSNR} & \textbf{SSIM}\\ 
    \hline
    \cite{exposure_fusion}'s quality score + \cite{exposure_fusion}'s RCS  & 15.14 & 0.652\\
    \cite{exposure_fusion}'s quality score + our RCS &  16.78 & 0.702 \\
    Our quality score + our RCS & 17.74 & 0.704\\
    \end{tabular}
    \caption{Comparison with the quality score and reference combination strategy (RCS) proposed in \cite{exposure_fusion}}
    \label{ablation_fusion_method}
\end{table}
\begin{figure*}[]
\centering
\includegraphics[width=\textwidth,page=14]{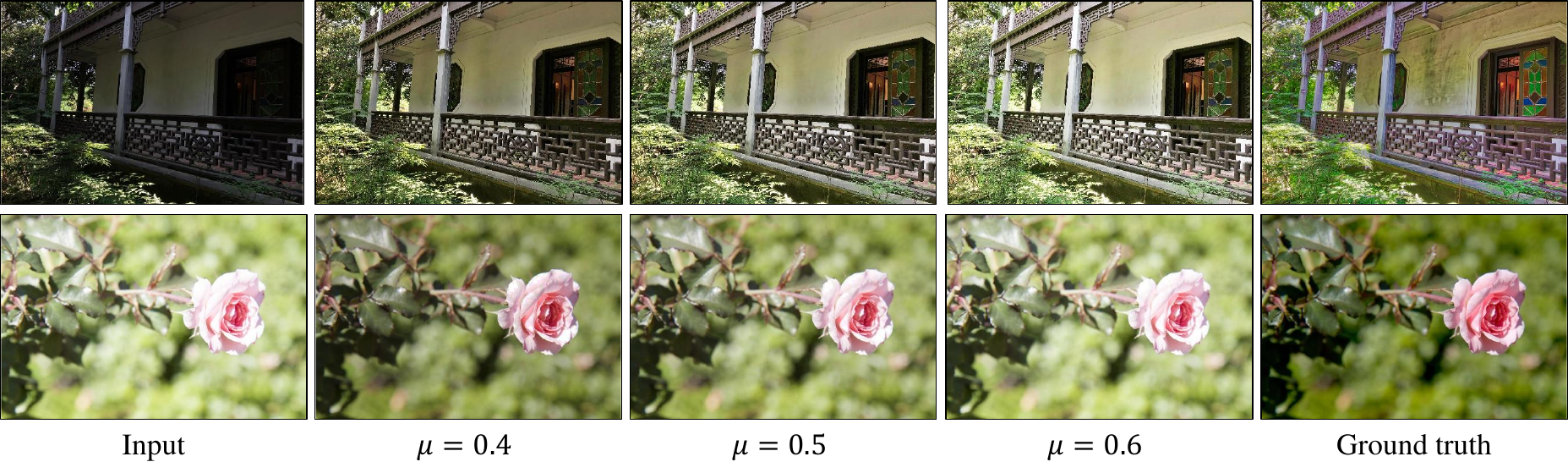}
\caption{Visual comparison among outputs of our model trained with different well-exposed level $\mu$. Training with a well-exposed level of 0.5 seems to balance our model in handling both under-exposed and over-exposed images}
\label{fig:ablation_exposedlevel}
\end{figure*}

\noindent\textbf{Well-exposed level.}
We conduct additional experiments to evaluate the effect of well-exposed level $\mu$ in the Equation (2) of our main paper on the performance of our approach. As shown in Figure~\ref{fig:ablation_exposedlevel} our model trained with a well-exposed level of 0.4  does not work effectively on under-exposed images while increasing this value to 0.6 makes our model fail to recover the detail of over-exposed images. Training with a well-exposed level of 0.5 seems to balance our model between these two cases, yielding the highest quantitative results, as shown in Table~\ref{tab:exposure_level}.

\begin{figure*}[]
\centering
\includegraphics[width=0.9\textwidth,page=4]{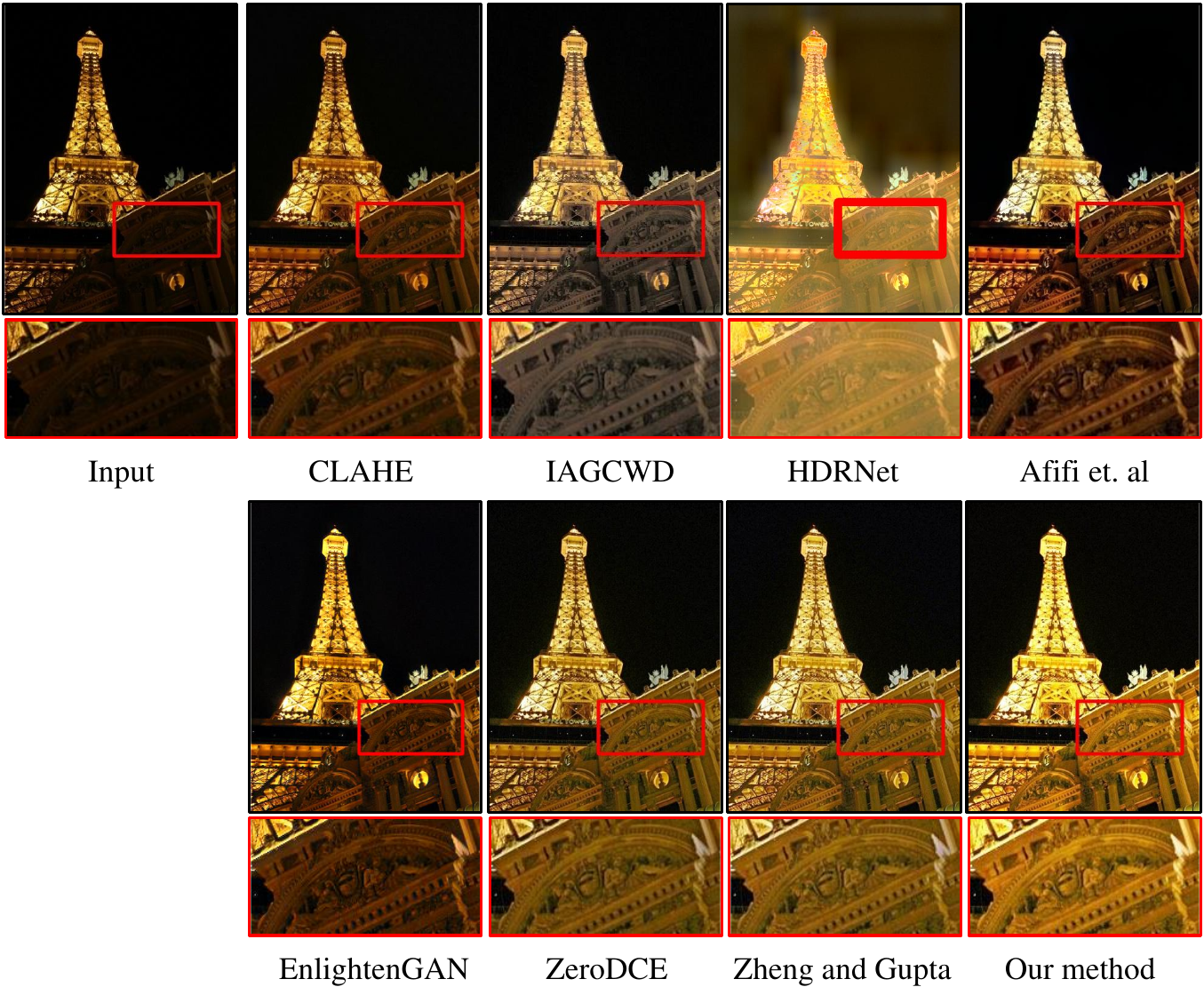}
\caption{Visual comparison on a lowlight image taken from the DICM dataset. The unsupervised methods including ZeroDCE \cite{zerodce}, Zheng and Gupta~\cite{semanticguided}, and our method produce more compelling results than others. Among them, our method's result is arguably the best in terms of contrast and color preservation as shown in boxed regions}
\label{fig:dicm1}
\end{figure*}
\begin{figure*}[]
\centering
\includegraphics[width=\textwidth,page=15]{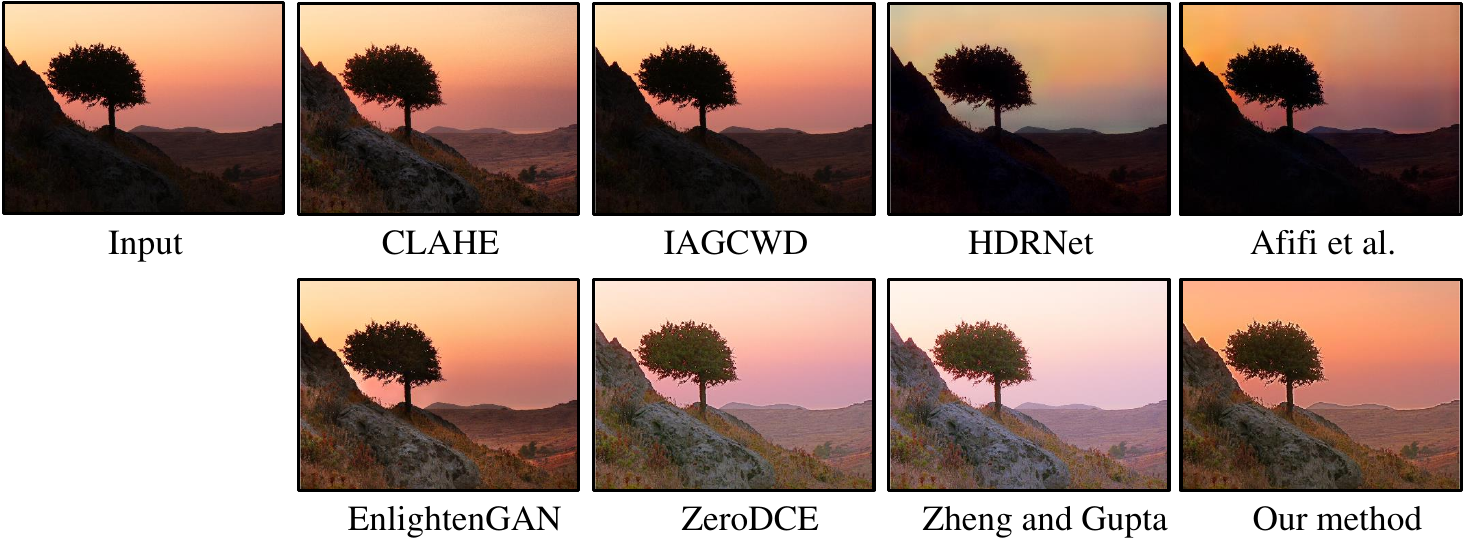}
\caption{Visual comparison on an image taken from the TMDIED dataset. Our result image seems to be more lively}
\label{fig:tmdied2}
\end{figure*}
\begin{figure*}[]
\centering
\includegraphics[width=\textwidth,page=18]{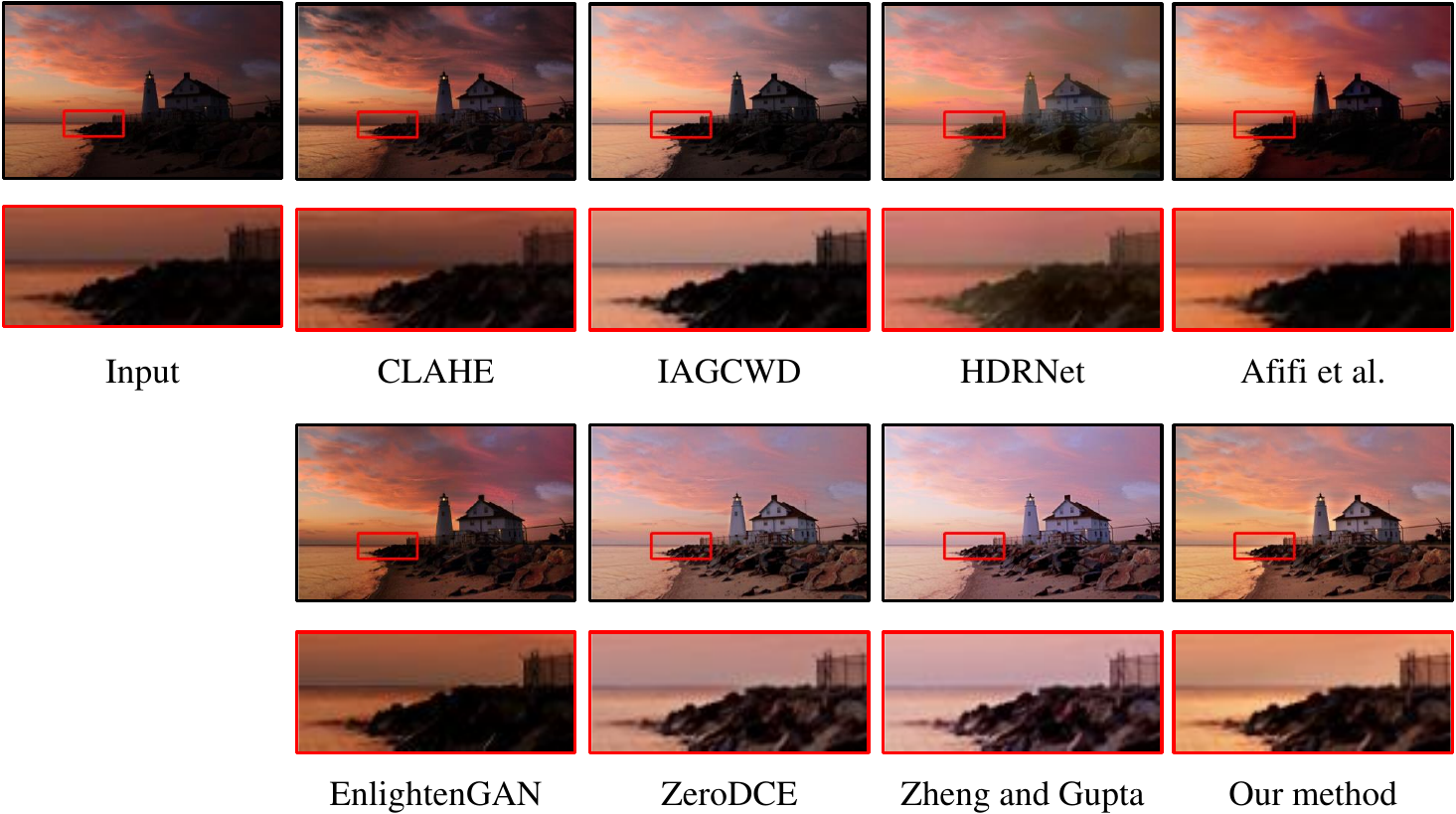}
\caption{Visual comparison on an image taken from the MEF dataset. Our model gives a better result in terms of enhancing under-exposed areas and preserving the original color temperature}
\label{fig:mef}
\end{figure*}
\begin{figure*}[]
\centering
\includegraphics[width=\textwidth,page=16]{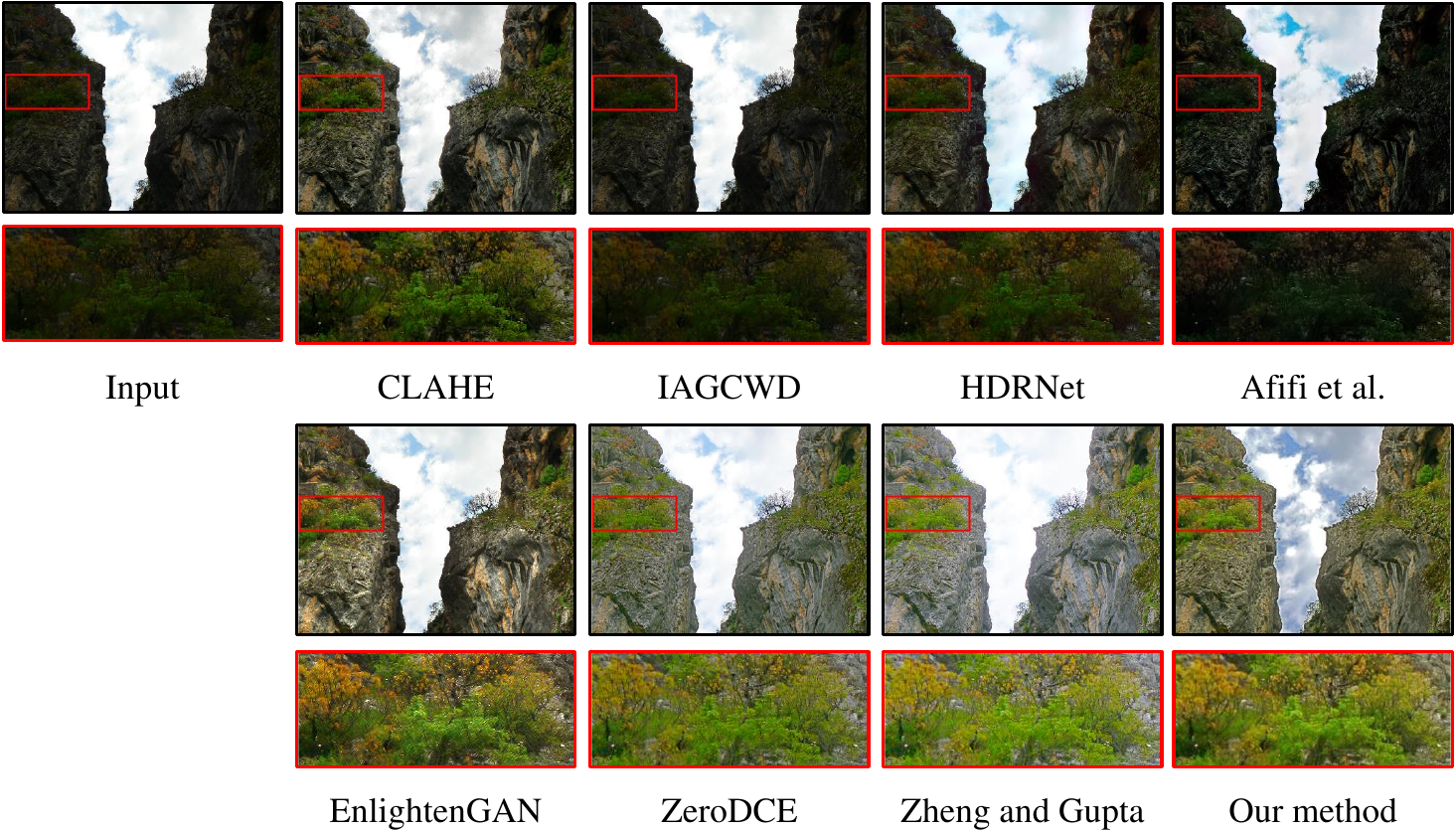}
\caption{Visual comparison on an image taken from the TMDIED dataset. Our method gives the best balance in contrast between the dark and the bright regions}
\label{fig:tmdied1}
\end{figure*}

\subsection{Visual Comparison Results} \label{supp_visualoutput}
This section presents additional qualitative results on other different public datasets including DICM~\cite{dicm}, MEF~\cite{mef}, TMDIED\footnote{https://sites.google.com/site/vonikakis/datasets}. We compare our method with two non-learning methods: CLAHE \cite{clahe}, IAGCWD \cite{IAGCWD},  an unpaired method EnlightenGAN \cite{enlightengan}, two unsupervised methods: ZeroDCE~\cite{zerodce}, Zheng and Gupta\cite{semanticguided}, and two supervised methods: HDRNet~\cite{hdrnet}, Afifi \etal{}~\cite{afifi2021learning}. The results are presented in Figures \ref{fig:dicm1}, \ref{fig:mef}, \ref{fig:tmdied1} and \ref{fig:tmdied2}. It is worth noting that all the learning-based methods are trained on the SICE dataset except the Afifi \etal{}~\cite{afifi2021learning} due to its Matlab license issue. 

{\small
\bibliographystyle{ieee_fullname}
\bibliography{egbib}
}

\end{document}